\documentclass[a4paper,orivec,runningheads]{llncs}
\usepackage{makeidx}  
\usepackage{amssymb}
\usepackage{url}     
\usepackage{amsmath}
\usepackage{nccmath}
\usepackage{amsfonts} 
\usepackage{graphicx}
\usepackage{multirow}
\usepackage{mathrsfs}
\graphicspath{{pics/}}
\usepackage{siunitx}
\usepackage{placeins}
\usepackage{esvect}
\usepackage{pbox}

\usepackage{color,soul}

\graphicspath{{pics/}{figs/}}

\usepackage{tikz}
\usetikzlibrary{arrows}
\usetikzlibrary{calc}
\usetikzlibrary{shapes.misc}
\pgfdeclarelayer{background}
\pgfdeclarelayer{foreground}
\pgfsetlayers{background,main,foreground}

\usepackage{color,soul}

\begin{document}
\frontmatter          
\mainmatter              
\title{Grey matter sublayer thickness estimation in the mouse cerebellum}%
\titlerunning{Grey matter sublayer thickness estimation in the mouse cerebellum}
\authorrunning{Da Ma\ \textit{et al.}}
\author{Da Ma\inst{1}, Manuel J. Cardoso\inst{1}, Maria A. Zuluaga\inst{1}, Marc Modat\inst{1}, Nick. Powell\inst{1}, Frances Wiseman\inst{3}, Victor Tybulewicz\inst{4}, Elizabeth Fisher\inst{3}, Mark. F. Lythgoe\inst{2}, S\'ebastien Ourselin\inst{1}}
\institute{Translational Imaging Group, CMIC, University College London, UK \and Centre for Advanced Biomedical Imaging , University College London, UK \and Institute of Neurology, University College London, UK, \and Crick Institute, UK} 
\maketitle
%
%
%
\begin{abstract}%
The cerebellar grey matter morphology is an important feature to study neurodegenerative diseases such as Alzheimer's disease or Down's syndrome. Its volume or thickness is commonly used as a surrogate imaging biomarker for such diseases.
Most studies about grey matter thickness estimation focused on the cortex, and little attention has been drawn on the morphology of the cerebellum.
Using \textit{ex vivo} high-resolution MRI, it is now possible to visualise the different cell layers in the mouse cerebellum.
In this work, we introduce a framework to extract the Purkinje layer within the grey matter, enabling the estimation of the thickness of the cerebellar grey matter, the granular layer and molecular layer from gadolinium-enhanced \textit{ex vivo} mouse brain MRI. Application to mouse model of Down's syndrome found reduced cortical and layer thicknesses in the transchromosomic group. 
\end{abstract}
%
%
%
\section{Introduction}
Grey matter thickness measurements have been widely used for quantitative analysis of neurodegenerative diseases. Studies have been limited to the cerebrum, even though this type of diseases also affect the cerebellum. Clinical studies have shown that diseases such as Down's Syndrome correlate with morphological changes in the cerebellar grey matter \cite{Aylward1997}. Although cerebellar grey matter thickness measurements could provide more insights about neurodegenerative diseases and phenotype identification, these are currently limited by the resolution of clinical MRI and the highly convoluted nature of the human brain.

Thanks to the high resolution of preclinical MRI scanners and to the anatomical similarities between human and mouse brains, the latter represents a promising model to further understand how the cerebellum is affected by the progression of neurodegenerative diseases. To date most of the quantitative analysis on mice cerebellum has focused on volumetric analysis and, to the best of our knowledge, no previous studies have addressed the problem of estimating the cerebellar cortical thickness.

In this work, we propose to not only estimate the thickness of the mouse cerebellar grey matter, but also extract and measure the thickness of two anatomical cerebellar layers (\textit{i.e.} the granular, and the molecular layer) from high resolution MRI. The presented framework explores and extends the Laplace equation based cortical thickness estimation method to multiple layers. These layers are defined through surface segmentation and, when not visible, extrapolated using the Laplace equation.



\section{Methods}
\begin{figure}[tb!]
	\centering
	\includegraphics[ width=0.9\textwidth]{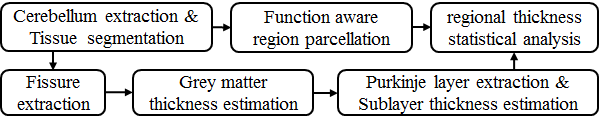}
	\caption{Schematic diagram of the proposed framework.}
	\label{Fig1}
  \end{figure}

In this section we introduce the proposed framework for cerebellum sublayer thickness estimation.
The overall pipeline is presented in Fig. \ref{Fig1}. The first steps of the framework include the extraction of the cerebellum, the white matter (WM) and grey matter (GM) tissue segmentation, the fissure extraction and the parcellation of the cerebellum based on the functional characteristics of the tissues. The later steps include the extraction of the Purkinje layer, which enables the estimation of sublayer thickness.

\subsection{Cerebellum extraction and tissue segmentation}

Based on a T2-weighted MR image, the cerebellum is extracted using a multi-atlas segmentation propagation framework~\cite{Ma2014}.
In order to  obtain an accurate thickness estimation, one needs first to obtain an accurate segmentation of the WM and GM tissues. Ashburner \& Friston based their approach on a Gaussian mixture model of tissue classes, and segmented the tissues within an expectation-maximisation framework ~\cite{Ashburner2000}. This framework gains in robustness by using a priori spatial information in the form of tissue probabilistic maps. Since there are currently no publicly available tissue probability maps for mouse cerebellum, we use a semi-automatic approach to first generate tissue probability maps, and then segment the tissues.

The intensity distribution in the cerebellar area for all MR images in the study is standardised using landmark-based piecewise linear intensity scaling introduced by Ny{\'u}l \textit{et al.}~\cite{Nyul2000} with 11 histogram landmarks.
Using an iterative groupwise scheme, we then create an average image. A symmetric block-matching approach~\cite{Modat2014} is used for the global registration steps and an symmetric scheme based on a cubic B-Spline parametrisation of a stationary velocity field~\cite{Modat2012a} is employed for the local registration stages.

Using a Gaussian mixture model of tissue classes, we first segment the group average image into 4 tissues without using anatomical priors. The segmentation of the group image is then manually corrected in order to generate an anatomically correct segmentation, e.g. disconnect misclassified Purkinje layer voxels from the white matter class. Note that this step only needs to be done once, as the groupwise tissue segmentation can be used to segment new subjects. 

Using the backward transformations obtained during the groupwise registration step, the segmented tissues are propagated back to the initial input images. As proposed by Chakravarty \textit{et al.}~\cite{Chakravarty2013}, all input images and their newly generated tissue segmentations are then considered as a template database for a multi-atlas segmentation propagation scheme. Using a leave-one-out framework, for each image, a spatial anatomical prior is generated by first propagating all the tissue segmentations from every other images to the target dataset, followed by a fusion step \cite{Ma2014}. Each image is then segmented using a Gaussian mixture model of tissue classes combined with the image specific probability maps.
\subsection{Fissure extraction}

The morphology of mouse cerebellum consists of folia lobes separated by fissures. Most of these thin fissures are subject to partial volume effect, resulting in incorrect tissue segmentations. In order to appropriately extract the fissures, even when highly corrupted by PV, a distance-based skeletonisation technique ~\cite{Han2004} is used here. A geodesic distance transform $D(x)$ is obtained by solving the Eikonal equation $F(x)|\nabla D(x)|=1$ from the white/grey matter boundary. Here, $D(x)$ is the geodesic distance and $F(x)$ is the speed function, in the computational domain $\Omega$, defined as $F=I\ast\mathcal{G}_{\sigma}$, with $I$ being the image and $\mathcal{G}_\sigma$ being a Gaussian kernel with $\sigma=1.5$ voxels. The fissures are then extracted by first finding the local maxima of $D(x)$ only along the direction of $\nabla D(x)$, followed by a recursive geodesic erosion in order to ensure single voxel thick fissure segmentations.
\subsection{Grey matter thickness estimation}
As proposed by Jones \textit{et al.}~\cite{Jones2000}, cortical thickness is estimated here by modelling the anatomical laminar layers as Laplacian field level-set analogues. The Laplace equation is solved using the Jacobi method. We first reconstruct the pial surface by combining the extracted fissures with the outer boundary of cerebellum mask. The thickness of each voxel is defined as the length of the streamlines perpendicularly passing through the isosurface of the Laplacian field. The streamlines are integrated along the direction of the normalised vector field $\hat{V}$, and are calculated using the Eulerian PDE method proposed by Yezzi and Prince~\cite{Yezzi2003}. At each voxel, the length of the streamlines along the direction $\hat{V}$, when measured from the pial surface is denoted by $D_\text{Pial}$, and when measured from the WM is denoted $D_{WM}$. The thickness at each voxel $x$ can thus be estimated by adding $D_{WM}(x)$ and $D_{\text{Pial}}(x)$, i.e. $T_{GM}(x)=D_{WM}(x)+D_{\text{Pial}}(x)$. 
\subsection{Purkinje layer extraction and sublayer thickness estimation}
In this step, we extract the narrow layer that sits in the middle of the grey matter - the Purkinje layer. This will allow us to obtain the thickness of the other two grey matter sublayers - the molecular layer and the granular layer. Segmentation is obtained by exploiting both the intensity information and the laminar nature of the Purkinje layer.

The Purkinje layer is an anatomical surface. Thus, a modified Frangi vesselness filter \cite{Frangi1998} is used here to find and enhance planar structures $\textit{P(s)}$ (instead of tubular ones) within the GM region at fixed scale $s=0.04$: 

$\textit{P}(s)=\begin{cases}
 0 & \text{ if } \lambda_2 > 0 \text{ or } \lambda_3 > 0 \\ 
 exp(-\frac{R_A^{2}}{2\alpha^{2}})
exp(-\frac{R_B^{2}}{2\beta^{2}})
(1-exp(-\frac{S^{2}}{2c^{2}}))
\end{cases}\text{ (1) },$

where $\alpha$,$\beta$ and $c$ are thresholds,  $R_A=\frac{|\lambda_2|}{|\lambda_3|}\text{ , } R_B=\frac{|\lambda_1|}{\sqrt{\left | \lambda_2\lambda_3 \right |}}$, and $\lambda_k$ are the k\textit{-th} smallest eigenvalue decomposition  ($|\lambda_1|\leqslant|\lambda_2|\leqslant|\lambda_3|$).

Areas with significant filter response are used as an initial estimates of the Purkinje layer $M_{P^0}$. Voxels touching the white matter or the sulcal region are removed through conditional morphological erosion.

\begin{figure}[tb!]
	\centering
	\includegraphics[width=\textwidth]{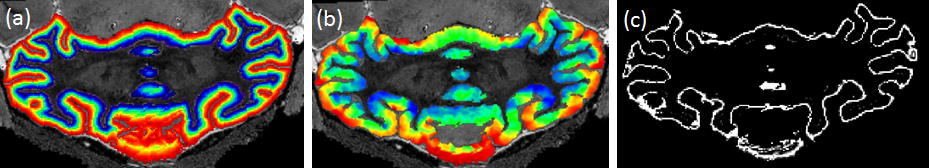}
	\caption{(a) White matter distance ratio map $R_{WM}$. (b) Gaussian smoothed map for the ratio map of Purkinje layer $R_{P_S}$. (c) Map of $\lambda$.}
	\label{Fig2}
\end{figure}

The above method does not capture the Purkinje in its entirety, mostly due to problems with local intensity contrast, partial volume effect, image noise, and adjacency to the grey matter boundary. Thus, the initial estimate of the Purkinje layer is used to extrapolate the remaining locations. In order to enhance the Purkinje layer segmentation, we first define the white matter thickness ratio as $R_{WM}=D_{WM}/T_{GM}$ (see Fig~\ref{Fig2}(a)), and the Purkinje layer distance ratio $R_{P}=R_{WM}*M_{P^0}$ at each grey matter voxel. 

A multi-level Gaussian smoothing is applied on $R_{P}$ to all voxel $x\in R_{WM} \cap x\not \in M_{P^0}$. 10 Gaussian smoothing levels with exponentially decreasing variances between 15 and 1 voxels were used in this work. This multi-level Gaussian smoothly propagates and averages the value of $R_P$ to neighbouring regions, providing an estimate of $R_P$ for voxels outside $M_{P^0}$. The smoothed version of $R_P$ is here denoted $R_{P_S}$ (Fig~\ref{Fig2}(b)). If the the Gaussian smoothed map $R_{P_S}$ is equal to $R_{WM}$ at location $x$, then the voxel $x$ should be part of the Purkinje layer. We thus define the  $\lambda=|R_{P_S}-R_{WM}|$ as a measurement of distance between the two maps (see Fig~\ref{Fig2}(c)).

In order to robustly find locations where $\lambda\approx 0$, we find local minima of $\lambda$ only along the direction of the vector field $\hat{V}$. The recovered local minima $\lambda_{min}$ are then added to the initial Purkinje layer mask, i.e. the final segmentation of the Purkinje layer is given by $M_{P^F}=M_{P^0}\cup\lambda_{min}$. 

Lastly, all relevant thicknesses, including whole grey matter thicknesses $T_{GM}$, granular layer thicknesses $T_{Gran}$, and molecular layer thickness $T_{Mol}$ are measured at the Purkinje layer location $M_{P^F}$. 
The granular layer thickness is here defined as the distance from the WM, given by $T(x)_{Gran}=D_{WM}(x)\ \forall x \in M_{P^F}$, the molecular layer thickness is defined as the distance from the pial layer, given by $T(x)_{Mol}=D_{Pial}(x)\ \forall x \in M_{P^F}$, and the total GM thickness at the Purkinje layer voxels, denoted by $T_{GM_{P^F}}$, is given by $T_{GM_{P^F}}(x)=T_{GM}(x) \forall x\in M_{P^F}$.

\subsection{Grey matter function aware region parcellation}

\begin{figure}[t!]
	\centering
	\includegraphics[width=\textwidth]{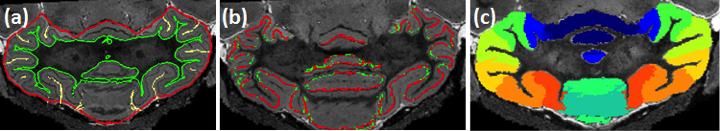}
	\caption{(a) Results of preprocessing steps. Red: cerebellum extraction; Green: White matter extraction; Yellow: Fissure extraction. (b) Purkinje layer extraction. Red: initial extraction; Green: final extraction. (c) Parcellated grey matter function regions.}
	\label{Fig_preprocessing}
\end{figure} 


Similarly to studies on human cortical thickness~\cite{Grand2013}, we measure the average thickness ($T_{GM}$, $T_{Gran}$ and $T_{Mol}$) in regions of interest. The grey matter sub-regions are parcellated automatically using the high resolution mouse cerebellum atlas database published by Ullmann \textit{et al.}~\cite{Ullmann2012}. This atlas divides the cerebellum into multiple regions based on their neuronal function. As previously, we use the approach by Chakravarty \textit{et al.}~\cite{Chakravarty2013} to obtain image specific parcellations based on a leave-on-out segmentation propagation scheme.

\section{Experimental data and validation}

\begin{figure}[tb!]
	\centering
	\includegraphics[width=\textwidth]{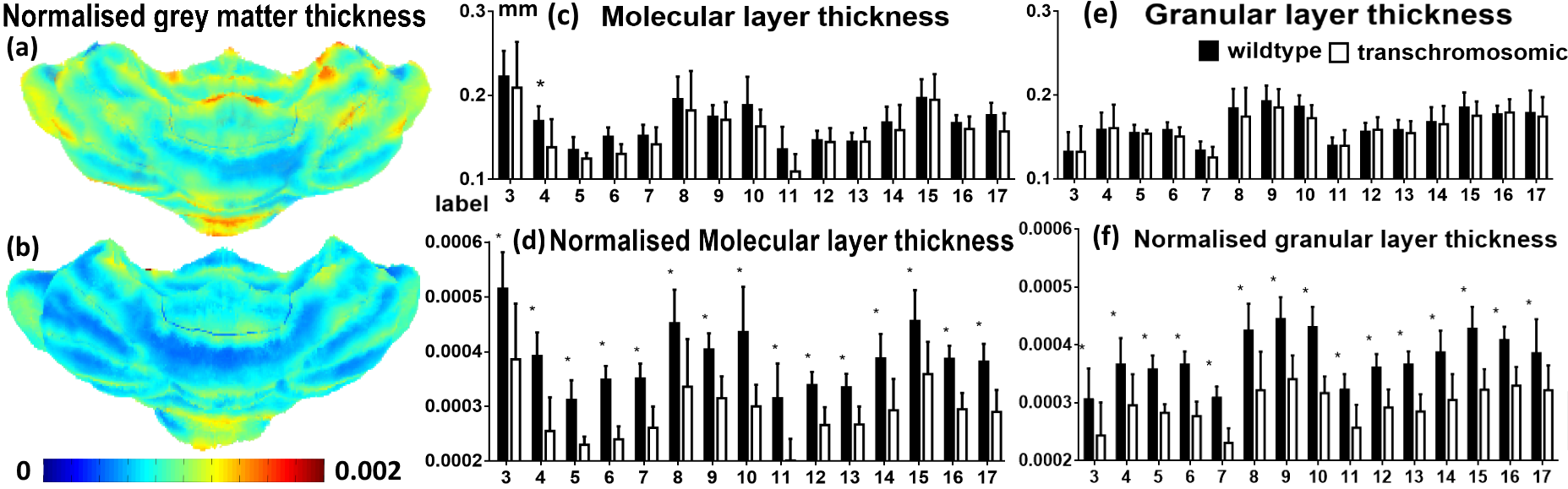}
	\caption{Normalised grey matter thickness map of (a) wildtype group, and (b)transgenic group, as well as the thicknesses of parcellated regions of (c,d) molecular layer, (e,f) granular layer before and after normalisation with TIV.  *: Significant difference between the wildtype and the transchromosomic group.}
	\label{Fig_Regional_Thickness}
\end{figure} 

The proposed method has been assessed on a data set which includes 28 gadolinium-enhanced \textit{ex vivo} T2 MRI scan of mouse brains: 14 wildtype and 14 transchromosomic that model Down's syndrome ~\cite{ODoherty2005a}. All theses images have been processed as previously described. Fig.~\ref{Fig_preprocessing}(a) shows an example of an extracted cerebellar grey matter. Fig.~\ref{Fig_preprocessing}(b) shows the initial and final Purkinje layer extraction. Fig.~\ref{Fig_preprocessing}(c) shows the parcellated cerebellar structures. 

We compare the full grey matter thickness as well as the sublayer thicknesses between the wild type and the transchromosomic groups for each parcellated region. To regress out the effect of the gross brain size, we also normalised the thickness to the total intracranial volume (TIV).  The results are presented in Fig.~\ref{Fig_Regional_Thickness}. We corrected for multiple comparisons with a false discovery rate set to $q=0.1$. Both the granular and molecular layer of the transchromosomic group are thinner over all the subregions when compared with the wild type group. Significant differences were found in a subregion of the molecular layer. After normalising with TIV, the difference becomes significant over all subregions for both sublayers.  Our results agree with previous findings  by Baxter \textit{et al.}, in which measurements from histological slides also showed thickness reduction of granular layer and molecular layer in mouse model of Down's syndrome\cite{Baxter2000}.

\section{Conclusion \& Discussion}

We presented a framework for mouse cerebellar grey matter sublayer thickness estimation. Sublayer measurements were enabled by a 2-step extraction of the Purkinje layer. The framework has been evaluated on \textit{ex vivo} MRI data of a mouse model of Down's syndrome, where data suggests a reduction in regional average thickness in mice that model Down's Syndrome.

\begin{figure}[tb!]
	\centering
	\includegraphics[width=0.8\textwidth]{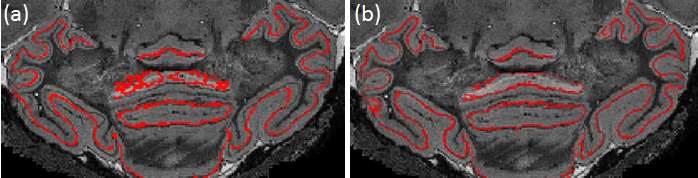}
	\caption{Comparison between the (a) central surface and (b) the Purkinje layer.}
	\label{Fig_Central_Surface}
\end{figure}

Within most current studies, the grey matter thickness estimation is commonly projected onto the central surface in order to perform quantitative statistical analyses~\cite{Han2004}. However, this model is limited by the spatial variability in sublayer thickness within the grey matter (see in Fig.~\ref{Fig_Central_Surface}). The information of the Purkinje layer location could provide robust information consistently across subjects, when performing groupwise analyses. Fig \ref{Fig_Central_Surface} shows the difference between the central surface and the Purkinje layer.

\begin{figure}[tb!]
	\centering
	\includegraphics[width=0.8\textwidth]{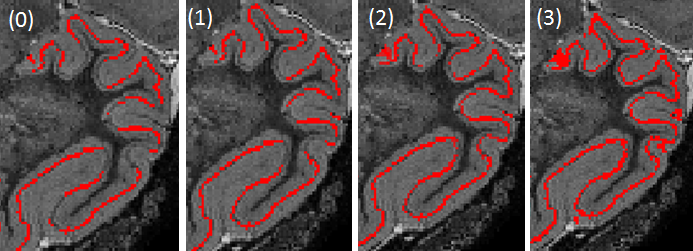}
	\caption{Iterative multi-kernel Gaussian smoothness gradually recover the Purkinje layer (0) Initial extracted Purkinje layer. (1-3) Recovered Purkinje layer after each iteration.}
	\label{Fig_Iteration}
\end{figure}

The fissure (sulcal) extraction is an important preprocessing step for cerebellar (cortical) thickness estimation. Previous studies have been using tissue membership functions to guide the skeletisation when extracting deep sulcal lines ~\cite{Han2004}. While the proposed method can greatly improve the Purkinje layer segmentation accuracy (see Fig \ref{Fig_Iteration}), due to complex intensity patterns and the presence of inter-layer contrast in the mouse cerebellum, an accurate tissue/Purkinje segmentation remains challenging. Future work will explore a joint model for the segmentation of the cerebellar tissues and Purkinje layer.

\begin{figure}[b!]
	\centering
	\includegraphics[width=\textwidth]{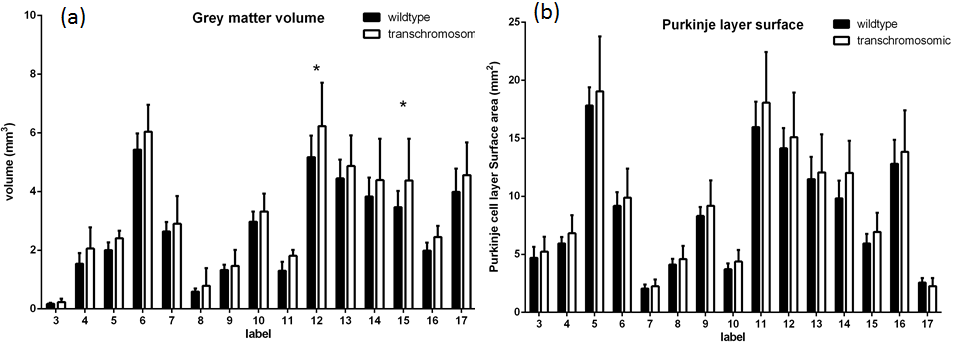}
	\caption{(a) Volume of the parcellated grey matter subregions. (b) Surface area of Purkinje layer.}
	\label{Fig_Volume_and_surface}
\end{figure}

To understand why the grey matter of the transchromosomic mouse is overall thinner, we also measured the volume of each parcellated grey matter region and the surface area of the Purkinje layer. The obtained results are shown in Fig.~\ref{Fig_Volume_and_surface}.
The larger volume and longer surface area of the Purkinje layer in the mice that model Down's Syndrome suggests that the morphology of the cerebellar grey matter of the transcromosomic mice could be more convoluted than the one of the  wild type mice. Further investigation is necessary to confirm these findings.

\bibliographystyle{splncs}
\bibliography{paper648}

\begin{thebibliography}{10}

\bibitem{Aylward1997}
Aylward, E.H., Habbak, R., Warren, A.C., Pulsifer, M.B., Barta, P.E., Jerram,
  M., Pearlson, G.D.:
\newblock Cerebellar volume in adults with down syndrome.
\newblock Archives of neurology \textbf{54}(2) (1997)  209--212

\bibitem{Ma2014}
Ma, D., Cardoso, M.J., Modat, M., Powell, N., Wells, J., Holmes, H., Wiseman,
  F., Tybulewicz, V., Fisher, E., Lythgoe, M.F.,  et~al.:
\newblock Automatic structural parcellation of mouse brain {MRI} using
  multi-atlas label fusion.
\newblock PLoS ONE \textbf{9}(1) (2014)

\bibitem{Ashburner2000}
Ashburner, J., Friston, K.J.:
\newblock Voxel-based morphometry—the methods.
\newblock Neuroimage \textbf{11}(6) (2000)  805--821

\bibitem{Nyul2000}
Ny{\'u}l, L.G., Udupa, J.K., Zhang, X.:
\newblock New variants of a method of mri scale standardization.
\newblock Medical Imaging, IEEE Transactions on \textbf{19}(2) (2000)  143--150

\bibitem{Modat2014}
Modat, M., Cash, D.M., Daga, P., Winston, G.P., Duncan, J.S., Ourselin, S.:
\newblock Global image registration using a symmetric block-matching approach.
\newblock Journal of Medical Imaging \textbf{1}(2) (2014)  024003

\bibitem{Modat2012a}
Modat, M., Daga, P., Cardoso, M., Ourselin, S., Ridgway, G., Ashburner, J.:
\newblock Parametric non-rigid registration using a stationary velocity field.
\newblock In: Mathematical Methods in Biomedical Image Analysis (MMBIA), 2012
  IEEE Workshop on. (2012)  145 --150

\bibitem{Chakravarty2013}
Chakravarty, M.M., Steadman, P., Eede, M.C., Calcott, R.D., Gu, V., Shaw, P.,
  Raznahan, A., Collins, D.L., Lerch, J.P.:
\newblock Performing label-fusion-based segmentation using multiple
  automatically generated templates.
\newblock Human brain mapping \textbf{34}(10) (2013)  2635--2654

\bibitem{Han2004}
Han, X., Pham, D.L., Tosun, D., Rettmann, M.E., Xu, C., Prince, J.L.:
\newblock Cruise: cortical reconstruction using implicit surface evolution.
\newblock NeuroImage \textbf{23}(3) (2004)  997--1012

\bibitem{Jones2000}
Jones, S.E., Buchbinder, B.R., Aharon, I.:
\newblock Three-dimensional mapping of cortical thickness using laplace's
  equation.
\newblock Human brain mapping \textbf{11}(1) (2000)  12--32

\bibitem{Yezzi2003}
Yezzi, A.J., Prince, J.L.:
\newblock An {E}ulerian {PDE} approach for computing tissue thickness.
\newblock Medical Imaging, IEEE Transactions on \textbf{22}(10) (2003)
  1332--1339

\bibitem{Frangi1998}
Frangi, A.F., Niessen, W.J., Vincken, K.L., Viergever, M.A.:
\newblock {Multiscale vessel enhancement filtering}.
\newblock In: Medial Image Computing and Computer-Assisted Invervention.
  Lecture Notes in Computer Science, vol 1496. Volume 1496. (1998)  130--137

\bibitem{Grand2013}
Grand'Maison, M., Zehntner, S.P., Ho, M.K., H{\'e}bert, F., Wood, A.,
  Carbonell, F., Zijdenbos, A.P., Hamel, E., Bedell, B.J.:
\newblock Early cortical thickness changes predict $\beta$-amyloid deposition
  in a mouse model of {A}lzheimer's disease.
\newblock Neurobiology of disease \textbf{54} (2013)  59--67

\bibitem{Ullmann2012}
Ullmann, J.F., Keller, M.D., Watson, C., Janke, A.L., Kurniawan, N.D., Yang,
  Z., Richards, K., Paxinos, G., Egan, G.F., Petrou, S.,  et~al.:
\newblock Segmentation of the c57bl/6j mouse cerebellum in magnetic resonance
  images.
\newblock Neuroimage \textbf{62}(3) (2012)  1408--1414

\bibitem{ODoherty2005a}
O'Doherty, A., Ruf, S., Mulligan, C., Hildreth, V., Errington, M.L., Cooke, S.,
  Sesay, A., Modino, S., Vanes, L., Hernandez, D., Linehan, J.M., Sharpe, P.T.,
  Brandner, S., Bliss, T.V.P., Henderson, D.J., Nizetic, D., Tybulewicz,
  V.L.J., Fisher, E.M.C.:
\newblock An aneuploid mouse strain carrying human chromosome 21 with down
  syndrome phenotypes.
\newblock Science \textbf{309}(5743) (2005)  2033--2037

\bibitem{Baxter2000}
Baxter, L.L., Moran, T.H., Richtsmeier, J.T., Troncoso, J., Reeves, R.H.:
\newblock {Discovery and genetic localization of Down syndrome cerebellar
  phenotypes using the Ts65Dn mouse.}
\newblock Human molecular genetics \textbf{9}(2) (2000)  195--202

\end{thebibliography}

\end{document}